\documentclass[10pt,twocolumn,letterpaper]{article}

\usepackage{cvpr}
\usepackage{times}
\usepackage{epsfig}
\usepackage{graphicx}
\usepackage{amsmath}
\usepackage{amssymb}
\usepackage{bm}
\usepackage{enumerate}
\usepackage{multirow}
\usepackage{array}

\usepackage{color}
\usepackage{caption}
\usepackage[normalem]{ulem}

\usepackage[pagebackref=true,breaklinks=true,letterpaper=true,colorlinks,bookmarks=false]{hyperref}

\cvprfinalcopy % *** Uncomment this line for the final submission

\def\cvprPaperID{4000} % *** Enter the CVPR Paper ID here

\ifcvprfinal\fi

\DeclareMathOperator*{\argmax}{arg\,max}

\begin{document}

%%%%%%%%% TITLE
\title{Incorporating External Knowledge to Answer Open-Domain Visual Questions \\ with Dynamic Memory Networks}

\author{Guohao Li, Hang Su, Wenwu Zhu\\
% Tsinghua National Laboratory for Information Science and Technology\\
Department of Computer Science and Technology, Tsinghua University, Beijing, China\\
{\tt\small ligh16@mails.tsinghua.edu.cn, suhangss@mail.tsinghua.edu.cn, wwzhu@tsinghua.edu.cn}
% \and
}

\maketitle
%\thispagestyle{empty}

%%%%%%%%% ABSTRACT
\begin{abstract}
 Visual Question Answering (VQA) has attracted much attention since it offers insight into the relationships between the multi-modal analysis of images and natural language. Most of the current algorithms are incapable of answering open-domain questions that require to perform reasoning beyond the image contents. To address this issue, we propose a novel framework which endows the model capabilities in answering more complex questions by leveraging massive external knowledge with dynamic memory networks. Specifically, the questions along with the corresponding images trigger a process to retrieve the relevant information in external knowledge bases, which are embedded into a continuous vector space by preserving the entity-relation structures. Afterwards, we employ dynamic memory networks to attend to the large body of facts in the knowledge graph and images, and then perform reasoning over these facts to generate corresponding answers. Extensive experiments demonstrate that our model not only achieves the state-of-the-art performance in the visual question answering task, but can also answer open-domain questions effectively by leveraging the external knowledge.

 % \hangx{Knowledge-Incorporated Dynamic Memory Network}
\end{abstract}

\section{Introduction}

Visual Question Answering (VQA) is a ladder towards a better understanding of the visual world, which pushes forward the
boundaries of both computer vision and natural language processing. A system in VQA tasks is given a text-based question about an image, which is expected to generate a correct answer corresponding to the question. In general, VQA is a kind of Visual Turing Test, which rigorously assesses whether a system is able to achieve human-level semantic analysis of images~\cite{geman2015visual,kafle2017visual}. A system could solve most of the tasks in computer vision if it performs as well as or better than humans in VQA. In this case, it has garnered increasing attentions due to its numerous potential applications~\cite{antol2015vqa}, such as providing a more natural way to improve human-computer interaction, enabling the visually impaired individuals to get information about images, etc.

\begin{figure}[!t]
  \begin{center}
    \includegraphics[width=1\linewidth]{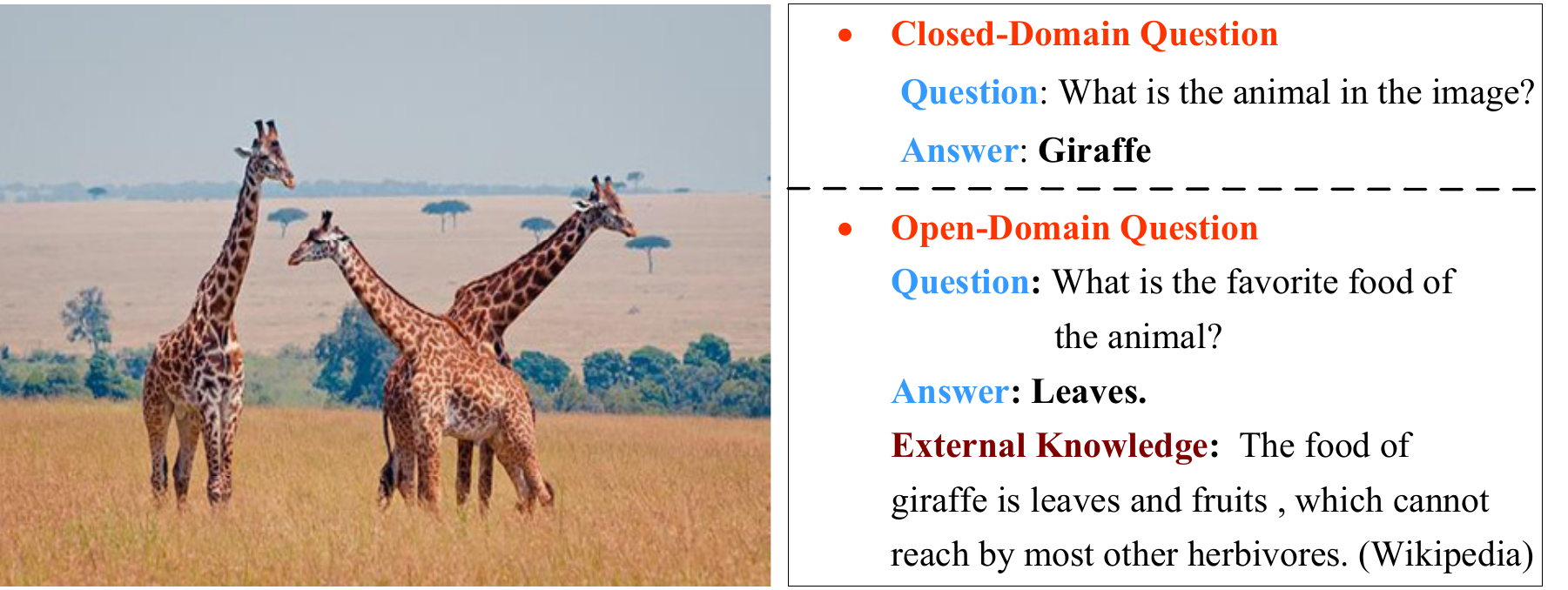}
  \end{center}
    \captionsetup{font={footnotesize}}
  \caption{A real case of open-domain visual question answering based on internal representation of an image and external knowledge. Recent success of deep learning provides a good opportunity to implement the closed-domain VQAs, but it is incapable of answering open-domain questions when external knowledge is needed. In this example, the system should recognize the giraffes and then query the knowledge bases for the main diet of giraffes. In this paper, we propose to explore the external knowledge along with the image representation based on a dynamic memory network, which allows a multi-hop reasoning over several facts.}
  \vspace{-3ex}
  \label{fig:demostration}
\end{figure}

To fulfill VQA tasks, it requires to endow the responder to understand intention of the question, reason over visual elements of the image, and sometimes have general knowledge about the world. Most of the present methods solve VQA by jointly learning interactions and performing inference over the question and image contents based on the recent success of deep learning~\cite{malinowski2015ask,antol2015vqa,ren2015image,gao2015you,fukui2016multimodal}, which can be further improved by introducing the attention mechanisms~\cite{zhu2016cvpr,yang2016stacked,xu2016ask,lu2016hierarchical,yu2017multi,anderson2017bottom}. However, most of questions in the current VQA dataset are quite simple, which are answerable by analyzing the question and image alone~\cite{antol2015vqa,wu2017visual}. It can be debated whether the system can answer questions that require prior knowledge ranging common sense to subject-specific and even expert-level knowledge.
It is attractive to develop methods that are capable of deeper image understanding by answering open-domain questions~\cite{wu2017visual}, which requires the system to have the mechanisms in connecting VQA with structured knowledge, as is shown in Fig.~\ref{fig:demostration}. Some efforts have been made in this direction, but most of them can only handle a limited number of predefined types of questions~\cite{wang2015explicit,wang2017fvqa}.

Different from the text-based QA problem, it is unfavourable to conduct the open-domain VQA based on the knowledge-based reasoning, since it is inevitably incomplete to describe an image with structured forms~\cite{krizhevsky2012imagenet}. The recent availability of large training datasets~\cite{wu2017visual} makes it feasible to train a complex model in an end-to-end fashion by leveraging the recent advances in deep neural networks (DNN)~\cite{antol2015vqa,gao2015you,zhu2016cvpr,lu2016hierarchical,anderson2017bottom}. Nevertheless, it is non-trivial to integrate knowledge into DNN-based methods, since the knowledge are usually represented in a symbol-based or graph-based manner (e.g., Freebase~\cite{bollacker2008freebase}, DBPedia~\cite{auer2007dbpedia}), which is intrinsically different from the DNN-based features. A few attempts are made in this direction~\cite{wu2016ask}, but it may involve much irrelevant information and fail to implement multi-hop reasoning over several facts.

The memory networks~\cite{weston2014memory,sukhbaatar2015end,kumar2016ask} offer an opportunity to address these challenges by reading from and writing to the external memory module, which is modeled by the actions of neural networks. Recently, it has demonstrated the state-of-the-art performance in numerous NLP applications, including the reading comprehension~\cite{pan2017memen} and textual question answering~\cite{bordes2015large,kumar2016ask}. Some seminal efforts are also made to implement VQA based on dynamic memory networks~\cite{xiong2016dynamic}, but it does not involve the mechanism to incorporate the external knowledge, making it incapable of answering open-domain visual questions. Nevertheless, the attractive characteristics motivate us to leverage the memory structures to encode the large-scale structured knowledge and fuse it with the image features, which offers an approach to answer open domain visual questions.

\begin{figure*}
  \begin{center}
  	\includegraphics[width=0.92\linewidth]{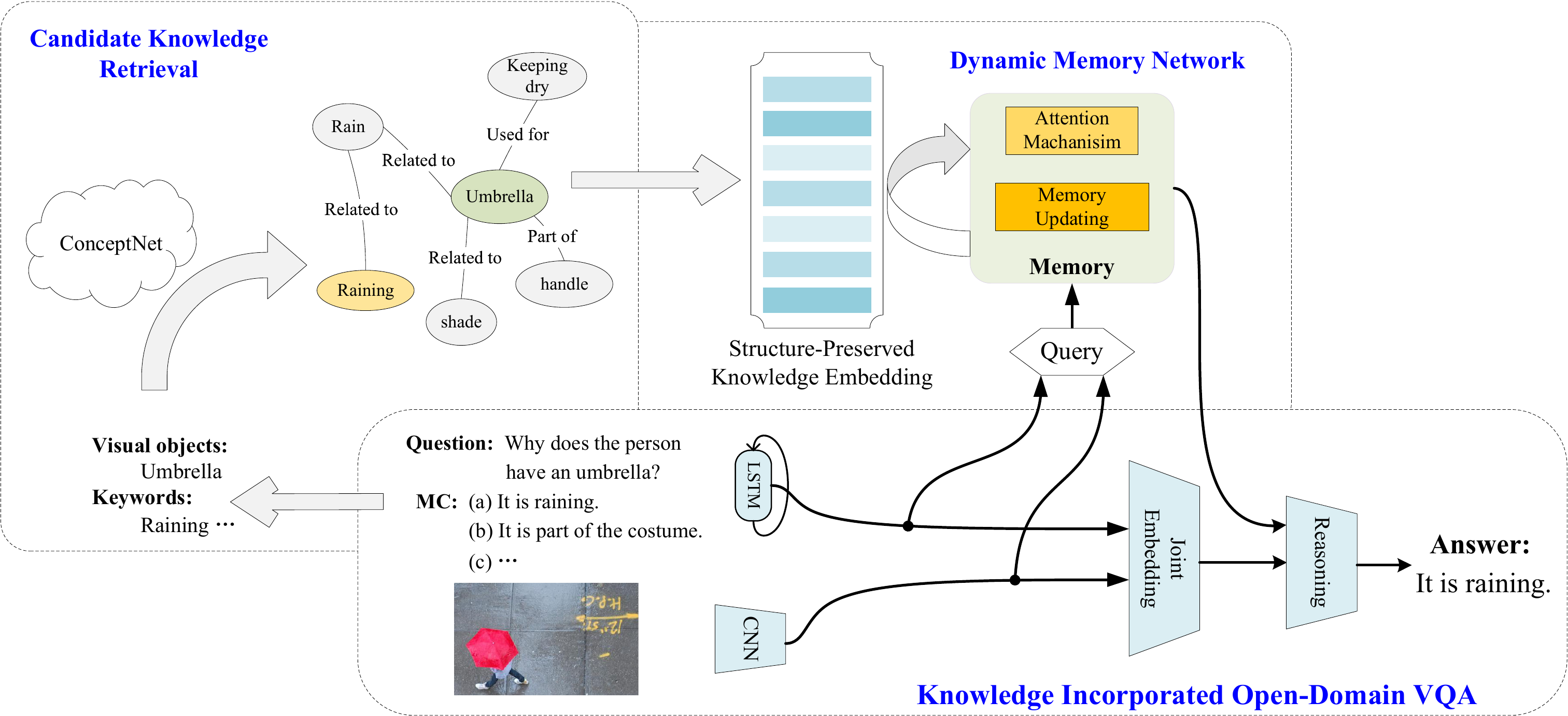}
  \end{center}
    \captionsetup{font={footnotesize}}
    \vspace{-2ex}
  \caption{Overall architecture of our proposed KDMN network. Given an image and the corresponding questions, the visual objects of the input image and key words of the corresponding questions are extracted using the Fast-RCNN and syntax analysis, respectively. Afterwards, we propose to assess the importance of entities in the knowledge graph and retrieve the most informative context-relevant knowledge triples, which are fed to the memory network after embedding the candidate knowledge into a continuous feature space. Consequentially, we integrate the representations of images and extracted knowledge into a common space, and store the features in a dynamic memory module. The open-domain VQA can be implemented by interpreting the joint representation under attention mechanism.}
  \vspace{-2ex}
  \label{fig:overview}
\end{figure*}

  % useful parts among these candidate knowledges.

\subsection{Our Proposal}
To address the aforementioned issues, we propose a novel Knowledge-incorporated Dynamic Memory Network framework (KDMN), which allows to introduce the massive external knowledge to answer open-domain visual questions by exploiting the dynamic memory network. It endows a system with an capability to answer a broad class of open-domain questions by reasoning over the image content incorporating the massive knowledge, which is conducted by the memory structures. 

Different from most of existing techniques that focus on answering visual questions solely on the image content, we propose to address a more challenging scenario which requires to implement reasoning beyond the image content. The DNN-based approaches~\cite{antol2015vqa,gao2015you,zhu2016cvpr} are therefore not sufficient, since they can only capture information present in the training images. Recent advances witness several attempts to link the knowledge to VQA methods~\cite{wang2015explicit, wang2017fvqa}, which make use of structured knowledge graphs and reason about an image on the supporting facts. Most of these algorithms first extract the visual concepts from a given image, and implement reasoning over the structured knowledge bases explicitly. However, it is non-trivial to extract sufficient visual attributes, since an image lacks the structure and grammatical rules as language. To address this issue, we propose to retrieve a bath of candidate knowledge corresponding to the given image and related questions, and feed them to the deep neural network implicitly. The proposed approach provides a general pipeline that simultaneously preserves the advantages of DNN-based approaches~\cite{antol2015vqa,gao2015you,zhu2016cvpr} and knowledge-based techniques~\cite{wang2015explicit,wang2017fvqa}.

In general, the underlying symbolic nature of a Knowledge Graph (KG) makes it difficult to integrate with DNNs. The usual knowledge graph embedding models such as TransE~\cite{bordes2013translating} focus on link prediction, which is different from VQA task aiming to fuse knowledge. To tackle this issue, we propose to embed the entities and relations of a KG into a continuous vector space, such that the factual knowledge can be used in a more simple manner. Each knowledge triple is treated as a three-word SVO $(subject, verb, object)$ phrase, and embedded into a feature space by feeding its word-embedding through an RNN architecture. In this case, the proposed knowledge embedding feature shares a common space with other textual elements (questions and answers), which provides an additional advantage to integrate them more easily.

Once the massive external knowledge is integrated into the model, it is imperative to provide a flexible mechanism to store a richer representation. The memory network, which contains scalable memory with a learning component to read from and write to it, allows complex reasoning by modeling interaction between multiple
parts of the data~\cite{weston2014memory,xiong2016dynamic}. In this paper, we adopt the most recent advance of Improved Dynamic Memory Networks (DMN+)~\cite{xiong2016dynamic} to implement the complex reasoning over several facts. Our model provides a mechanism to attend to candidate knowledge embedding in an iterative manner, and fuse it with the multi-modal data including image, text and knowledge triples in the memory component. The memory vector therefore memorizes useful knowledge to facilitate the prediction of the final answer.
Compared with the DMN+~\cite{xiong2016dynamic}, we introduce the external knowledge into the memory network, and endows the system an ability to answer open-domain question accordingly.

To summarize, our framework is capable of conducting the multi-modal data reasoning including the image content and external knowledge, such that the system is endowed with a more general capability of image interpretation. Our main contributions are as follows:
\begin{itemize}
% \vspace{-0.5ex}
\item To our best knowledge, this is the first attempt to integrating the external knowledge and image representation with a memory mechanism, such that the open-domain visual question answering can be conducted effectively with the massive knowledge appropriately harnessed;
\item We propose a novel structure-preserved method to embed the knowledge triples into a common space with other textual data, making it flexible to integrate different modalities of data in an implicit manner such as  image, text and knowledge triples;

 \item We propose to exploit the dynamic memory network to implement multi-hop reasonings, which has a capability to automatically retrieve the relevant information in the knowledge bases and infer the most probable answers accordingly.
\end{itemize}

\section{Overview}

In this section, we outline our model to implement the open-domain visual question answering. In order to conduct the task, we propose to incorporate the image content and external knowledge by exploiting the most recent advance of dynamic memory network~\cite{kumar2016ask, xiong2016dynamic}, yielding three main modules in Fig.~\ref{fig:overview}. The system is therefore endowed with an ability to answer arbitrary questions corresponding to a specific image.

Considering of the fact that most of existing VQA datasets include a minority of questions that require prior knowledge, the performance therefore cannot reflect the particular capabilities. We automatically produce a collection of more challenging question-answer pairs, which require complex reasoning beyond the image contents by incorporating the external knowledge. We hope that it can serve as a benchmark for evaluating the capability of various VQA models on the open-domain scenarios
% \footnote{All the data will be released after the double-blind review}
.

Given an image, we apply the Fast-RCNN~\cite{girshick2015fast} to detect the visual objects of the input image, and extract keywords of the corresponding questions with syntax analysis. Based on these information, we propose to learn a mechanism to retrieve the candidate knowledge by querying the large-scale knowledge graph, yielding a subgraph of relevant knowledge to facilitate the question answering. During the past years, a substantial amount of large-scale knowledge bases have been developed, which store common sense and factual knowledge in a machine readable fashion. In general, each piece of structured knowledge is represented as a triple $(subject, rel, object)$ with $subject$ and $object$ being two entities or concepts, and $rel$ corresponding to the specific relationship between them. In this paper, we adopt external knowledge mined from ConceptNet~\cite{speer2012conceptnet}, an open multilingual knowledge graph containing common-sense relationships between daily words, to aid the reasoning of open-domain VQA.

Our VQA model provides a novel mechanism to integrate image information with that extracted from the ConceptNet within a dynamic memory network. In general, it is non-trivial to integrate the structured knowledge with the DNN features due to their different modalities. To address this issue, we embed the entities and relations of the subgraph into a continuous vector space, which preserves the inherent structure of the KGs. The feature embedding provides a convenience to fuse with the image representation in a dynamic memory network, which builds on the attention mechanism and the memory update mechanism. The attention mechanism is responsible to produce the contextual vector with relevance inferred by the question and previous memory status. The memory update mechanism then renews the memory status based on the contextual vector, which can memorize useful information for predicting the final answer. The novelty lies the fact that these disparate forms of information are embedded into a common space based on memory network, which facilities the subsequent answer reasoning.

Finally, we generate a predicted answer by reasoning over the facts in the memory along with the image contents. In this paper, we focus on the task of multi-choice setting, where several multi-choice candidate answers are provided along with a question and a corresponding image. For each question, we treat every multi-choice answer as input, and predict whether the image-question-answer triplet is correct. The proposed model tries to choose one candidate answer with the highest probability by inferring the cross entropy error on the answers through the entire network.

\section{Answer Open-Domain Visual Questions}

In this section, we elaborate on the details and formulations of our proposed model
for answering open-domain visual questions. We first retrieve an appropriate amount of candidate knowledge from the large-scale ConceptNet by analyzing the image content and the corresponding questions; afterward, we propose a novel framework based on dynamic memory network to embed these symbolic knowledge triples into a continuous vector space and store it in a memory bank; finally, we exploit these information to implement the open-domain VQA by fusing the knowledge with image representation.

\subsection{Candidate Knowledge Retrieval }

In order to answer the open-domain visual questions, we should sometime access information not present in the image by retrieving the candidate knowledge in the KBs. A desirable knowledge retrieval should include most of the useful information while ignore the irrelevant ones, which is essential to avoid model misleading and reduce the computation cost. To this end, we take the following three principles in consideration as (1) entities appeared in images and questions (\textit{key entities}) are critical; (2) the importance of entities that have direct or indirect links to \textit{key entities} decays as the number of link hops increases; (3) edges between these entities are potentially useful knowledge.

Following these principles, we propose a three-step procedure to retrieve that candidate knowledge that are relevant to the context of images and questions. The retrieval procedure pays more attention on graph nodes that are linked to semantic entities, which also takes account of graph structure for measuring edge importance.

In order to retrieve the most informative knowledge, we first extract the candidate nodes in the ConceptNet by analyzing the prominent visual objects in images with Fast-RCNN~\cite{girshick2015fast}, and textual keywords with the Natural Language Toolkit~\cite{bird2009natural}. Both of them are then associated with the corresponding semantic entities in ConceptNet~\cite{speer2012conceptnet} by matching all possible n-grams of words. Afterwards, we retrieve the first-order subgraph using these selected nodes from ConceptNet~\cite{speer2012conceptnet}, which includes all edges connecting with at least one candidate node. It is assumed that the resultant subgraph contains the most relevant information, which is sufficient to answer questions by reducing the redundancy. The resultant first-order knowledge subgraph is denoted as $G$.

Finally, we compress the subgraph $G$ by evaluating and ranking the importance of edges in $G$ using a designed score function, and carefully select the top-$N$ edges along with the nodes for subsequent task. Specifically, we first assign initial weights $w_{i}$ for each subgraph node, e.g., the initial weights for visual object can be proportional to their corresponding bounding-box area such that the dominant objects receive more attention, the textual keywords are treated equally. Then, we calculate the importance score of each node in $G$ by traversing each edge and propagating node weights to their neighbors with a decay factor $r\in(0,1)$ as
\begin{align}
  score(i)=w_{i}+\sum_{j \in G \backslash i} r ^n w_{j},
\end{align}
where $n$ is the number of link hops between the entity $i$ and $j$.
For simplicity, we ignore the edge direction and edge type (relation type), and define the importance of edge $w_{i,j}$ as the weights sum of two connected nodes as
\begin{equation}
  w_{i,j}=score(i)+score(j), \quad \forall (i,j) \in G.
\end{equation}
In this paper, we take the top-$N$ edges ranked by $w_{i,j}$ as the final candidate knowledge for the given context, denoted as $G^\ast$.

\subsection{Knowledge Embedding in Memories}

The candidate knowledge that we have extracted is represented in a symbolic triplet format, which is intrinsically incompatible with DNNs. This fact urges us to embed the entities and relation of knowledge triples into a continuous vector space. Moreover, we regard each \textit{entity-relation-entity} triple as one knowledge unit,
since each triple naturally represents one piece of fact. The knowledge units can be stored in memory slots for reading and writing, and distilled through an attention mechanism for the subsequent tasks.

In order to embed the symbolic knowledge triples into memory vector slots, we treat the entities and relations as words, and map them into a continuous vector space using word embedding~\cite{pennington2014glove}. Afterwards, the embedded knowledge is encoded into a fixed-size vector by feeding it to a recurrent neural network (RNN).
Specifically, we initialize the word-embedding matrix with a pre-trained GloVe word-embedding~\cite{pennington2014glove}, and refine it simultaneously with the rest of procedure of question and candidate answer embedding. In this case, the entities and relations share a common embedding space with other textual elements (questions and answers), which makes them much more flexible to fuse later.

Afterwards, the knowledge triples are treated as SVO phrases of $(subject, verb, object)$, and fed to to a standard two-layer stacked LSTM as
\begin{align}
  &C^{(t)}_{i} = \text{LSTM}\left(\mathbf{L}[w^{t}_{i}], C^{(t-1)}_{i}\right), \\
  & t=\{1,2,3\}, \text{ and  } i=1, \cdots, N,\nonumber
\end{align}
where $w^{t}_{i}$ is the $t_{\text{th}}$ word of the $i_{\text{th}}$ SVO phrase, $(w^{1}_{i},w^{2}_{i},w^{3}_{i}) \in G^\ast$,
$\mathbf{L}$ is the word embedding matrix~\cite{pennington2014glove}, and $C_{i}$ is the internal state of LSTM cell when forwarding the $i_{\text{th}}$ SVO phrase. The rationale lies in the fact that the LSTM can capture the semantic meanings effectively when the knowledge triples are treated as SVO phrases.

For each question-answering context, we take the LSTM internal states of the relevant knowledge triples as memory vectors, yielding the embedded knowledge stored in memory slots as
\begin{equation}
  \mathbf{M}=\left[C^{(3)}_{i}\right],
\end{equation}
where $\mathbf{M}(i)$ is the $i_{\text{th}}$ memory slot corresponding to the $i_{\text{th}}$ knowledge triples, which can be used for further answer inference. Note that the method is different from the usual knowledge graph embedding models, since our model aims to fuse knowledge with the latent features of images and text, whereas the alternative models such as TransE~\cite{bordes2013translating} focus on link prediction task.

\subsection{Attention-based Knowledge Fusion with DNNs}

We have stored $N$ relevant knowledge embeddings in memory slots for a given question-answer context, which allows to incorporate massive knowledge when $N$ is large.
The external knowledge overwhelms other contextual information in quantity, making it imperative to distill the useful information from the candidate knowledge.
The Dynamic Memory Network (DMN)~\cite{kumar2016ask, xiong2016dynamic} provides a mechanism to address the problem
by modeling interactions among multiple data channels.
In the DMN module, an episodic memory vector is formed and updated during an iterative attention process,
which memorizes the most useful information for question answering.
Moreover, the iterative process brings a potential capability of multi-hop reasoning.

This DMN consists of an attention component which generates a contextual vector using the previous memory vector, and an episodic memory updating component which updates itself based on the contextual vector. Specifically, we propose a novel method to generate the query vector $\mathbf{q}$ by feeding visual and textual features to a non-linear fully-connected layer to capture question-answer context information as
\begin{equation}
  \mathbf{q} = \tanh\left(\mathbf{W}_{1}
    \left[\mathbf{f}^{(I)};\mathbf{f}^{(Q)};\mathbf{f}^{(A)}\right]+\mathbf{b}_{1}\right),
\end{equation}
where $\mathbf{W}_1$ and $\mathbf{b}_{1}$ are the weight matrix and bias vector, respectively; and, $\mathbf{f}^{(I)}$, $\mathbf{f}^{(Q)}$ and $\mathbf{f}^{(A)}$ are denoted as DNN features corresponding to the images, questions and multi-choice answers, respectively.  The query vector $\mathbf{q}$ captures information from question-answer context.% \hangx{What is the physical meaning of $q$?}

During the training process, the query vector $\mathbf{q}$ initializes an episodic memory vector $\mathbf{m}^{(0)}$ as $\mathbf{m}^{(0)}=\mathbf{q}$. A iterative attention process is then triggered, which gradually refines the episodic memory $\mathbf{m}$ until the maximum number of iterations steps $T$ is reached. By the $T_{\text{th}}$ iteration, the episodic memory $\mathbf{m}^{(T)}$will memorize useful visual and external information to answer the question. % \hangx{Where do we arrive after the iteration?}

\textbf{Attention component}.
At the $t_{\text{th}}$ iteration, we concatenate each knowledge embedding $\mathbf{M}_{i}$ with last iteration episodic memory $\mathbf{m}^{(t-1)}$
and query vector $\mathbf{q}$,
then apply the basic soft attention procedure to obtain the $t_{\text{th}}$ context vector $\mathbf{c}^{(t)}$ as
\begin{align}
  \mathbf{z}_{i}^{(t)} &= \left[\mathbf{M}_{i};\mathbf{m}^{(t-1)};\mathbf{q}\right] \\
  \boldsymbol{\alpha}^{(t)} &= softmax\left(\mathbf{w}\tanh\left(\mathbf{W}_{2}\mathbf{z}_{i}^{(t)}+\mathbf{b}_{2}\right) \right) \\
  \mathbf{c}^{(t)}&=\sum_{i=1}^{N}\alpha_{i}^{(t)}\mathbf{M}_{i} \quad t=1, \cdots, T,
\end{align}
where $\mathbf{z}_{i}^{(t)}$ is the concatenated vector for the $i_{\text{th}}$ candidate memory at the $t_{\text{th}}$ iteration;
$\boldsymbol{\alpha}_{i}^{(t)}$ is the $i_{\text{th}}$ element of $\boldsymbol{\alpha}^{(t)}$ representing the normalized attention weight for $\mathbf{M}_{i}$ at the $t_{\text{th}}$ iteration; and,
$\mathbf{w}$, $\mathbf{W}_{2}$ and $\mathbf{b}_{2}$ are parameters to be optimized in deep neural networks. 

Hereby, we obtain the contextual vector $\mathbf{c}^{(t)}$, which captures useful external knowledge for updating episodic memory $\mathbf{m}^{(t-1)}$ and providing the supporting facts to answer the open-domain questions.
% The procedure is guided by $\mathbf{m}^{(t-1)}$ itself along with the query vector $\mathbf{q}$.

\textbf{Episodic memory updating component.}
We apply the memory update mechanism~\cite{sukhbaatar2015end, xiong2016dynamic} as
\begin{equation}
  \mathbf{m}^{(t)}=ReLU\left(\mathbf{W}_{3}
    \left[\mathbf{m}^{(t-1)};\mathbf{c}^{(t)};\mathbf{q}\right]+\mathbf{b}_{3}\right),
\end{equation}
where $\mathbf{W}_{3}$ and $\mathbf{b}_{3}$ are parameters to be optimized. After the iteration, the episodic memory $\mathbf{m}^{(T)}$ memorizes useful knowledge information to answer the open-domain question.

Compared with the DMN+ model implemented in ~\cite{xiong2016dynamic}, we allows the dynamic memory network to incorporate the massive external knowledge into procedure of VQA reasoning. It endows the system with the capability to answer more general visual questions relevant but beyond the image contents, which is more attractive in practical applications. 

\textbf{Fusion with episodic memory and inference.} Finally, we embed visual features $\mathbf{f}^{(I)}$ along with the textual features $\mathbf{f}^{(Q)}$ and $\mathbf{f}^{(A)}$ to a common space, and fuse them together using Hadamard product (element-wise multiplication) as
\begin{align}
  &\mathbf{e}^{(k)}=\tanh\left(\mathbf{W}^{(k)}\mathbf{f}^{(k)}+\mathbf{b}^{(k)}\right), k \in \{I, Q, A\} \\
  &\mathbf{h} =\mathbf{e}^{(I)} \odot \mathbf{e}^{(Q)} \odot \mathbf{e}^{(A)},
\end{align}
where $\mathbf{e}^{(I)}$, $\mathbf{e}^{(Q)}$ and $\mathbf{e}^{(A)}$ are embedded features for image, question and answer,
respectively; $\mathbf{h}$ is the fused feature in this common space; and,
$\mathbf{W}^{(I)}$, $\mathbf{W}^{(Q)}$ and $\mathbf{W}^{(A)}$ are corresponding to the parameters in neural networks.

The final episodic memory $\mathbf{m}^{(T)}$ are concatenated with the fused feature $\mathbf{h}$ to predict the probability of whether the multi-choice candidate answer is correct as
\begin{equation}
  ans^* = \argmax_{ans \in \{1,2,3,4\}}
  softmax\left(\mathbf{W}_{4}\left[\mathbf{h}_{ans};\mathbf{m}^{(T)}_{ans}\right]+\mathbf{b}_{4}\right),
\end{equation}
where $ans$ represents the index of multi-choice candidate answers; the supported knowledge triples are stored in $\mathbf{m}^{(T)}_{ans}$; and, $\mathbf{W}_{4}$ and $\mathbf{b}_{4}$ are the parameters to be optimized in the DNNs. The final choice are consequentially obtained once we have $ans^\ast$.
% where $ans$ represents the index of multi-choice candidate answers. $P_{ans}(true)$ is the probability of this candidate answer being $true$. $\mathbf{W}_{4}$ and $\mathbf{b}_{4}$ are the parameters to be optimized in the last classification layer of DNNs. The final choice are consequentially obtained once we have $ans^\ast$.

Our training objective is to learn parameters based on a cross-entropy loss function as
\begin{equation}
  \mathcal{L} = -\frac{1}{D}\sum_{i}^{D}\big(y_{i}\log \hat{y_{i}}+(1-y_{i})\log(1-\hat{y_{i}})\big),
\end{equation}
where
$\hat{y_{i}}=p_{i}(A^{(i)}|I^{(i)},Q^{(i)},K^{(i)};\boldsymbol{\theta})$ represents the probability of predicting the answer $A^{(i)}$, given the $i_{\text{th}}$ image $I^{(i)}$, question $Q^{(i)}$ and external knowledge $K^{(i)}$; $\boldsymbol{\theta}$ represents the model parameters; 
$D$ is the number of training samples; and $y_{i}$ is the label for the $i_{\text{th}}$ sample. The model can be trained in an end-to-end manner once we have the candidate knowledge triples are retrieved from the original knowledge graph.  

\section{Experiments}

In this section, we conduct extensive experiments to evaluate performance of our proposed model, and compare it with its variants and the alternative methods. We specifically implement the evaluation on a public benchmark dataset (Visual7W)~\cite{zhu2016cvpr} for the close-domain VQA task, and also generate numerous arbitrary question-answers pairs automatically to evaluate the performance on open-domain VQA. In this section, we first briefly review the dataset and the implementation details, and then report the performance of our proposed method comparing with several baseline models on both close-domain and open-domain VQA tasks.

\subsection{Datasets}
We train and evaluate our model on a public available large-scale visual question answering datasets,
the Visual7W dataset~\cite{zhu2016cvpr}, due to the diversity of question types.
Besides, since there is no public available open-domain VQA dataset for evaluation now,
we automatically build a collection of open-domain visual question-answer pairs
to examine the potentiality of our model for answering open-domain visual questions.

\subsubsection{Visual7W Dataset}
The Visual7W dataset~\cite{zhu2016cvpr} is built based on a subset of images from Visual Genome~\cite{krishna2017visual}, which includes questions in terms of \textit{(what, where, when, who, why, which and how)} along with the corresponding answers in a multi-choice format. Similar as~\cite{zhu2016cvpr}, we divide the dataset into training, validation and test subsets, with totally 327,939 question-answer pairs on 47,300 images. Compared with the alternative dataset, Visual7W has a diverse type of question-answer and image content~\cite{wu2017visual}, which provides more opportunities to assess the human-level capability of a system on the open-domain VQA.

\subsubsection{Open-domain Question Generation}
In this paper, we automatically generate numerous question-answer pairs by considering the image content and relevant background knowledge, which provides a test bed for the evaluation of a more realistic VQA task. Specifically, we generate a collection automatically based on the test image in the Visual7W by filling a set of question-answer templates, which means that the information is not present during the training stage. To make the task more challenging, we selectively sample the question-answer pairs that need to reasoning on both visual concept in the image and the external knowledge, making it resemble the scenario of the open-domain visual question answering. In this paper, we generate 16,850 open-domain question-answer pairs on images in Visual7W test split. More details on the QA generation and relevant information can be found in the supplementary material.
% All the data and source code will be released after the double-blind review, with a hope that it can serve as a benchmark for evaluating the capability of various VQA models on the open-domain scenarios.

\begin{figure*}[!t]
  \begin{center}
  	\includegraphics[width=0.92\linewidth]{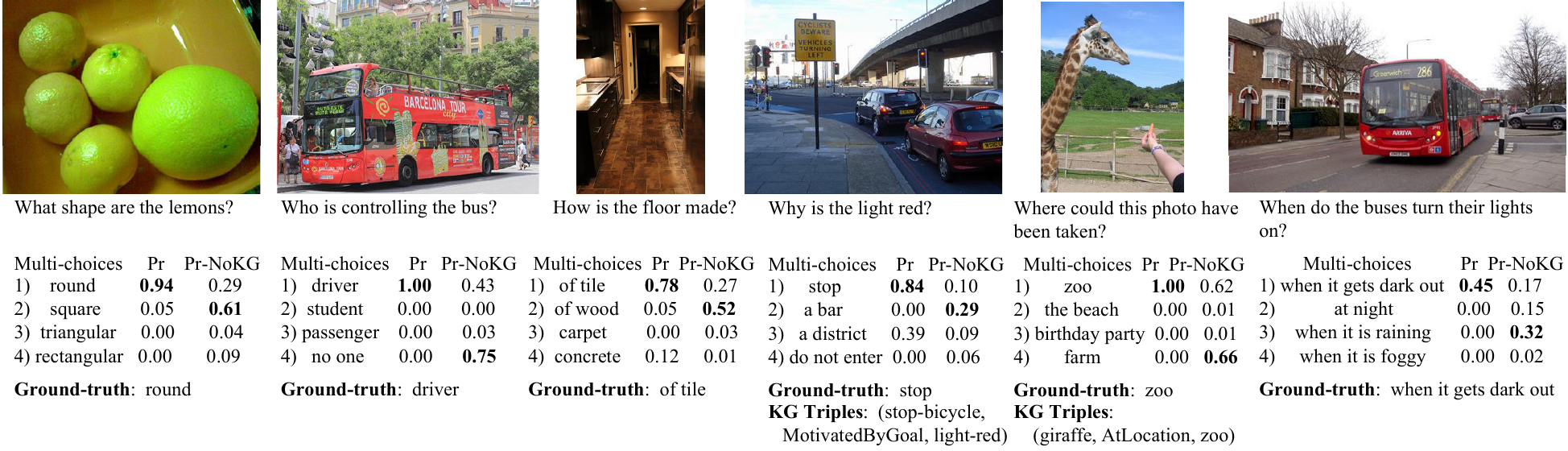}
  \end{center}
  \vspace{-1ex}
  \captionsetup{font={footnotesize}}
  \vspace{-2ex}
  \caption{Example results on the Visual7W dataset for (close-domain) VQA tasks. Given an image and the corresponding question, we report the corresponding answers obtained via our algorithm. Specifically, \textit{pr} denotes the predicted probability generated by our model, and \textit{pr-NoKG} is the predicted probability by the ablative model of \textit{KDMN-NoKG}. We make the predicted choices bold accordingly. The external knowledge triples are also provided if they are retrieved to support the joint reasoning by our method automatically. As is observed, the external knowledge is essential even for the conventional VQA tasks, e.g., in the 5th example, it is much easier to infer the place accordingly by incorporating external knowledge when a giraffe is recognized. 
  }
    \vspace{-2ex}
  \label{fig:samples-v7w}
\end{figure*}

\subsection{Implementation Details}
In our experiments,  we fix the joint-embedding common space dimension as 1024, word-embedding dimension as 300 and the dimension of LSTM internal states as 512. We use a pre-trained ResNet-101~\cite{he2016deep} model to extract image feature, and select 20 candidate knowledge triples for each QA pair through the experiments. Empirical study demonstrates it is sufficient in our task although more knowledge triples are also allowed. The iteration number of a dynamic memory network update is set to 2, and the dimension of episodic memory is set to 2048, which is equal to the dimension of memory slots.

In this paper, we combine the candidate Question-Answer pair to generate a hypothesis, and formulate the multi-choice VQA problem as a classification task. The correct answer can be determined by choosing the one with the largest probability. In each iteration, we randomly sample a batch of 500 QA pairs, and apply stochastic gradient descent algorithm with a base learning rate of 0.0001 to tune the model parameters. The candidate knowledge is first retrieved, and other modules are trained in an end-to-end manner.

\vspace{-1ex}
\subsubsection{Comparison Methods}
In order to analyze the contributions of each component in our knowledge-enhanced, memory-based model,
we ablate our full model as follows:
\begin{itemize}
\item \textit{KDMN-NoKG}: baseline version of our model. % \hangx{Rename base as DNN}
  No external knowledge involved in this model. Other parameters are set the same as full model.
\item \textit{KDMN-NoMem}: a version without memory network. External knowledge triples are used by one-pass soft attention.
\item \textit{KDMN}: our full model. External knowledge triples are incorporated in Dynamic Memory Network.
\end{itemize}

We also compare our method with several alternative VQA methods including (1) \textbf{LSTM-Att}~\cite{zhu2016cvpr}, a LSTM model with spatial attention; (2) \textbf{MemAUG}~\cite{ma2017visual}: a memory-augmented model for VQA; (3) \textbf{MCB+Att}~\cite{fukui2016multimodal}: a model combining multi-modal features by Multimodal Compact Bilinear pooling; (4) \textbf{MLAN}~\cite{yu2017multi}: an advanced multi-level attention model.
% \item MLP \cite{jabri2016revisiting}: a simple network with a two-layer MLP (multi-layer perceptron).
% \item MLP+query generator\cite{Zhu_2017_CVPR}: a model based on MLP model and iterative querying external information.\comment{which of them should I include?}

% \begin{table}
%   \begin{center}
%       \caption{Accuracy on our generated open-domain dataset}
%     \label{tab:opendomain}
%     \resizebox{0.55\columnwidth}{!}{
%       \begin{tabular}{|l|c|}
%         \hline
%         Methods & Accuracy \\
%         \hline\hline
%         base+glove & 58.55 \\
%         base+glove+kg20(ours) & 67.70 \\
%         base+glove+kg20+dmn(ours) & 78.07 \\
%         Base & 45.13 \\
%         Base + KG & 51.88 \\
%         Base + KG + DMN & 57.79 \\
%         Ensemble & \\
%         \hline
%       \end{tabular}
%     }
%   \end{center}
%   \vspace{-3ex}
% \end{table}

\begin{figure*}
  \begin{center}
  	\includegraphics[width=0.92\linewidth]{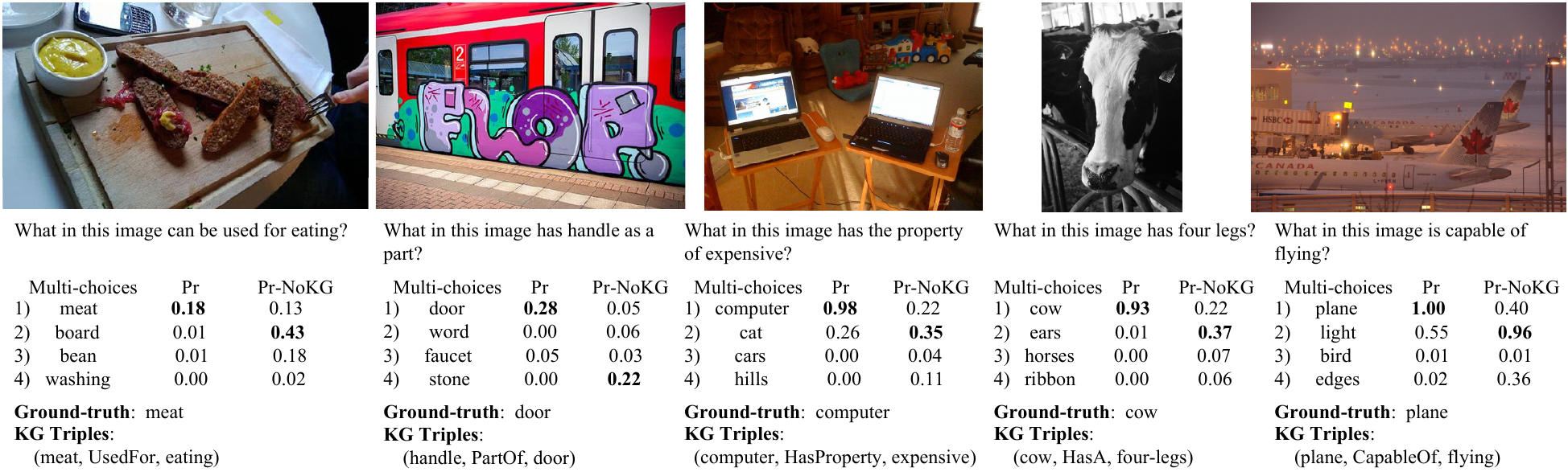}
  \end{center}
  \vspace{-2ex}
  \captionsetup{font={footnotesize}}
  \caption{Example results of open-domain visual question answering based on our proposed knowledge-incorporate dynamic memory network. Given an images, we automatically generate the open-domain question-answer pair by considering of the image content and the relevant background knowledge. We report the corresponding answers obtained via our algorithm. Specifically, \textit{pr} denotes the predicted probability generated by our model, and \textit{pr-NoKG} is the predicted probability by the ablative model of \textit{KDMN-NoKG}. The results demonstrate that external knowledge plays an essential role in answer open-questions. A system is incapable of inferring the food in the 1st example and the stuff prices in the 3rd example.}
  \vspace{-2ex}
  \label{fig:samples-opendomain}
\end{figure*}

\subsection{Results and Analysis}
In this section, we report the quantitative evaluation along with representative samples of our method, compared with our ablative models and the state-of-the-art method for both the conventional (close-domain) VQA task and open-domain VQA.

\subsubsection{VQA Task}

In this section, we report the quantitative accuracy in Table~\ref{tab:visual7w} along with the sample results in~\ref{fig:samples-v7w}. The overall results demonstrate that our algorithm obtains different boosts compared with the competitors on various kinds of questions, e.g., significant improvements on the questions of \textit{Who} ($5.9\%$), and \textit{What} ($4.9\%$) questions, and slightly boost on the questions of \textit{When} ($1.4\%$) and \textit{How} ($2.0\%$). After inspecting the success and failure cases, we found that the \textit{Who} and \textit{What} questions have larger diversity in questions and multi-choice answers compared to other types, therefore benefit more from external background knowledge. Note that compared with the method of MemAUG \cite{ma2017visual} in which a memory mechanism is also adopted, our algorithm still gain significant improvement, which further confirms our belief that the background knowledge provides critical supports.

We further make comprehensive comparisons among our ablative models. To make it fair, all the experiments are implemented on the same basic network structure and share the same hyper-parameters. In general, our \textit{KDMN} model on average gains $1.6\%$ over the \textit{KDMN-NoMem} model and $4.0\%$ over the \textit{KDMN-NoKG} model, which further implies the effectiveness of dynamic memory networks in exploiting external knowledge. Through iterative attention processes, the episodic memory vector captures background knowledge distilled from external knowledge embeddings. The \textit{KDMN-NoMem} model gains $2.4\%$ over the \textit{KDMN-NoKG} model, which implies that the incorporated external knowledge brings additional advantage, and act as a supplementary information for predicting the final answer. The indicative examples in Fig.~\ref{fig:samples-v7w} also demonstrate the impact of external knowledge, such as the 4th example of ``Why is the light red?''. It would be helpful if we could retrieve the function of the traffic lights from the external knowledge effectively.

\begin{table}[!bht]
  \vspace{-1ex}
  \begin{center}
    \caption{Accuracy on Visual7W dataset}
      \vspace{-1ex}
    \label{tab:visual7w}
    \resizebox{1\linewidth}{!}{
      \begin{tabular}{|l|c|c|c|c|c|c|c|}
        \hline
        Methods & What & Where & When & Who & Why & How & Average \\
        \hline\hline
        LSTM-Att.\cite{zhu2016cvpr} & 51.5 & 57.0 & 75.0 & 59.5 & 55.5 & 49.8 & 54.3 \\
        % Ma \etal \cite{ma2017visual} & 59.0 & 63.2 & 75.7 & 60.3 & 56.2 & 52.0 & 59.4 \\
        MCB + Att.\cite{fukui2016multimodal} & 60.3 & 70.4 & 79.5 & 69.2 & 58.2 & 51.1 & 62.2 \\
        MemAUG \cite{ma2017visual} & 62.2 & 68.9 & 76.8 & 66.4 & 57.8 & 52.9 & 62.8 \\
        MLAN \cite{yu2017multi} & 60.5 & 71.2 & 79.6 & 69.4 & 58.0 & 50.8 & 62.4 \\
        % MLP \cite{jabri2016revisiting} & 64.5 & 75.9 & 82.1 & 72.9 & 68.0 & 56.4 & 67.1 \\
        % MLP + query generator\cite{Zhu_2017_CVPR} & 65.1 & 77.8 & 80.7 & 81.4 & 65.3 & 54.1 & 67.9 \\
        \hline\hline
        KDMN-NoKG & 59.7 & 69.6 & 79.9 & 68.0 & 61.6 & 51.3 & 62.0 \\
        KDMN-NoMem & 62.1 & 71.5 & 81.1 & 72.5 & 62.9 & 54.0 & 64.4 \\
        KDMN & 64.6 & 73.1 & 81.3 & 73.9 & 64.1 & 53.3 & 66.0 \\
        Ensemble & 67.9 & 77.0 & 83.3 & 77.2 & 69.0 & 56.8 & 69.4 \\
        \hline
      \end{tabular}
    }
  \end{center}
  \vspace{-6ex}
\end{table}
% We find that our single\comment{try ensemble to reduce variance} \textit{Base+KG+DMN} model
% can correctly answer most of the questions (over $80\%$) that are correctly answered by \textit{Base} model,
% and a large portion questions (almost $40\%$) that can not correctly answered by \textit{Base} model.
% These facts indicate that our model preserves a sound ability answering close-domain questions with additional advantages for the questions.

\subsubsection{Open-Domain VQA}
In this section, we report the quantitative performance of open-domain VQA in Table~\ref{tab:opendomain} along with the sample results in Fig.~\ref{fig:samples-opendomain}. Since most of the alternative methods do not provide the results in the open-domain scenario, we make comprehensive comparison with our ablative models. As expected, we observe that a significant improvement ($12.7\%$) of our full \textit{KDMN} model over the \textit{KDMN-NoKG} model, where $6.8\%$ attributes to the involvement of external knowledge and $5.9\%$ attributes to the usage of memory network. Examples in Fig.~\ref{fig:samples-opendomain} further provide some intuitive understanding of our algorithm. It is difficult or even impossible for a system to answer the open domain question when comprehensive reasoning beyond image content is required, e.g., the background knowledge for prices of stuff is essential for a machine when inferring the expensive ones. The larger performance improvement on open-domain dataset supports our belief that background knowledge is essential to answer general visual questions. Note that the performance can be further improved if the technique of ensemble is allowed. We fused the results of several KDMN models which are trained from different initializations. Experiments demonstrate that we can further obtain an improvement about $3.1\%$.

\begin{table}[!h]
  \begin{center}
      \caption{Accuracy on our generated open-domain dataset. }
    \label{tab:opendomain}
    \vspace{-1ex}
    \resizebox{0.5\linewidth}{!}{
      \begin{tabular}{|l|c|}
        \hline
        Methods & Accuracy \\
        \hline\hline
        KDMN-NoKG & 45.1 \\
        KDMN-NoMem & 51.9 \\
        KDMN & 57.8 \\
        Ensemble & 60.9 \\
        \hline
      \end{tabular}
    }
  \end{center}
  \vspace{-5ex}
\end{table}

\section{Conclusion}
In this paper, we proposed a novel framework named knowledge-incorporate dynamic memory network (KDMN) to answer open-domain visual questions by harnessing massive external knowledge in dynamic memory network. Context-relevant external knowledge triples are retrieved and embedded into memory slots, then distilled through a dynamic memory network to jointly inference final answer with visual features. The proposed pipeline not only maintains the superiority of DNN-based methods, but also acquires the ability to exploit external knowledge for answering open-domain visual questions. Extensive experiments demonstrate that our method achieves competitive results on public large-scale dataset,
and gain huge improvement on our generated open-domain dataset.

{\small
\bibliographystyle{ieee}
\bibliography{references}
}

\onecolumn
\section{Supplementary Material}
\subsection{Details of our Open-domain Dataset Generation}

\begin{table*}[!b]
  \vspace{-2ex}
  \begin{center}
    \resizebox{0.6\linewidth}{!}{
      \begin{tabular}{|c|c|}
        \hline
        KG Triple & QA templates \\
        \hline\hline
        (\textit{visual}, UsedFor, \textit{other}) & what in this image can be used for \textit{\{other\}}? \\
        (\textit{other}, UsedFor, \textit{visual}) & what in this image can \textit{\{other\}} be used for? \\
        \hline
        (\textit{visual}, PartOf, \textit{other}) & what in this image is a part of \textit{\{other\}}? \\
        (\textit{other}, PartOf, \textit{visual}) & what in this image has \textit{\{other\}} as a part?? \\
        \hline
        (\textit{visual}, HasProperty, \textit{other}) & what in this image has the property of \textit{\{other\}}? \\
        (\textit{other}, HasProperty, \textit{visual}) & what property does the \textit{\{other\}} in this image have? \\
        \hline
        (\textit{visual}, HasA, \textit{other}) & what in this image has \textit{\{other\}}? \\
        (\textit{other}, HasA, \textit{visual}) & what in this image belongs to \textit{\{other\}}? \\
        \hline
        (\textit{visual}, CapableOf, \textit{other}) & what in this image is capable of \textit{\{other\}}? \\
        (\textit{other}, CapableOf, \textit{visual}) & what in this image is \textit{\{other\}} capable of? \\
        \hline
      \end{tabular}
    }
  \end{center}
  \captionsetup{font={footnotesize}}
  \vspace{-2ex}
  \caption{Templates for generate open-domain question-answer pairs.
    \textit{\{visual\}} is the KG entity representing visual object.
    \textit{\{other\}} is the KG entity that has a connection with \textit{\{visual\}}.
    We take \textit{\{visual\}} as the generated ground-truth answer.
  }
  \label{tab:templates}
\end{table*}

We obey several principles when building the open-domain VQA dataset for evaluation:
(1) The question-answer pairs should be generated automatically;
(2) Both of visual information and external knowledge should be required when answering these generated open-domain visual questions;
(3) The dataset should in multi-choices setting, in accordance with the Visual7W dataset for fair comparison.

The open-domain question-answer pairs are generated based on a subset of images in Visual7W~\cite{zhu2016cvpr} standard test split,
so that the test images are not present during the training stage.
For one particular image that we need to generate open-domain question-answer pairs about,
we firstly extract several prominent visual objects
and randomly select one visual object.
After linked to a semantic entity in ConceptNet~\cite{speer2012conceptnet},
the visual object connects other entities in ConceptNet through various relations, e.g. \textit{UsedFor, CapableOf},
and forms amount of knowledge triples $(head, relation, tail)$, where either $head$ or $tail$ is the visual object.
Again, we randomly select one knowledge triple,
and fill into a $relation$-related question-answer template to obtain the question-answer pair.
These templates assume that the correct answer satisfies knowledge requirement as well as appear in the image,
as shown in table~\ref{tab:templates}.

\begin{figure*}[!b]
  \begin{center}
  	\includegraphics[width=0.92\linewidth]{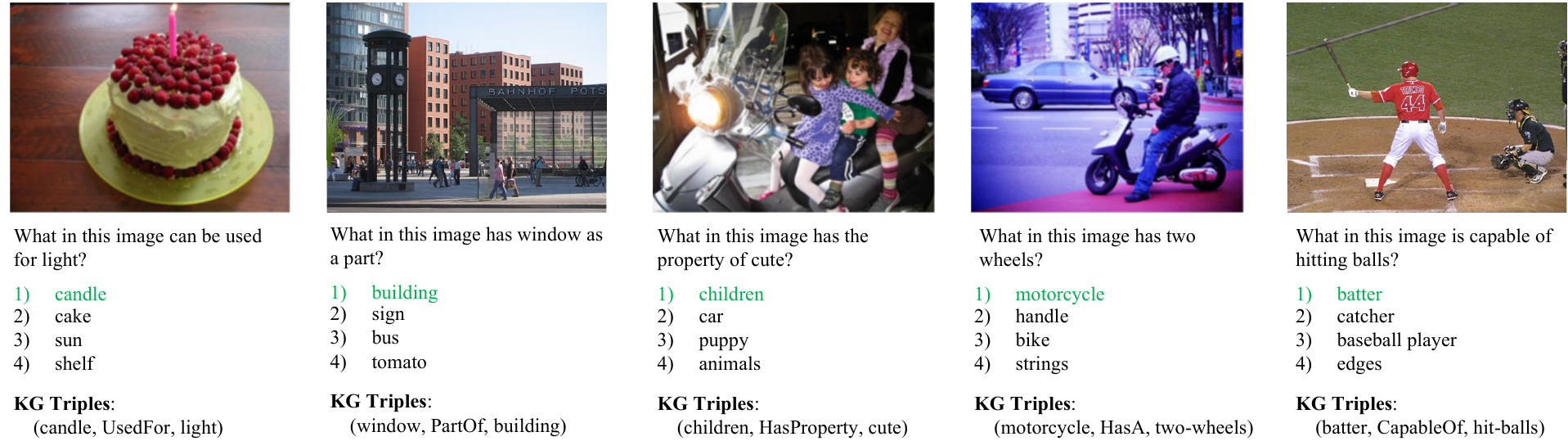}
  \end{center}
  \captionsetup{font={footnotesize}}
  \vspace{-2ex}
  \caption{Examples from our generated open-domain dataset.
    We mark ground-truth answers green.
    The bottom KG triples just provide insights into the generation process,
    and will not be included in the dataset.
    The candidate answers can be quite confusing in some questions, e.g., in the 1st example,
    the ground-truth ``candle'' appearing in the image can be used for light,
    while ``cake'' also appears in the image but cannot be used for light,
    ``sun'' can also be used for light but not appear in the image.
  }
  \label{fig:examples}
\end{figure*}

For each open-domain question-answer pair, we generate three additional confusing items as candidate answers.
These candidate answers are randomly sampled from a collection of answers,
which is composed of answers from other question-answer pairs belonging to the same $relation$ type.
In order to make the open-domain dataset more challenging,
we selectively sample confusing answers,
which either satisfy knowledge requirement or appear in the image,
but not satisfy both of them as the ground-truth answers do.
Specifically, one of the confusing answers satisfies knowledge requirement but not appears in image,
so that the model must attend to visual objects in image;
another one of the confusing answers appears in the image but not satisfies knowledge requirement,
so that the model must reason on external knowledge to answer these open-domain questions.
Please see examples in Figure~\ref{fig:examples}.

In total, we generate 16,850 open-domain question-answer pairs based on 8,425 images in Visual7W test split.

\end{document}